\documentclass{article}

\PassOptionsToPackage{numbers, compress}{natbib}

\usepackage[final]{neurips_2025}

\usepackage[utf8]{inputenc} 
\usepackage[T1]{fontenc}    
\usepackage{hyperref}       
\usepackage{url}            
\usepackage{booktabs}       
\usepackage{amsfonts}       
\usepackage{nicefrac}       
\usepackage{microtype}      
\usepackage{xpatch}
\usepackage{xcolor}
\usepackage{graphicx}
\usepackage{float}
\usepackage{booktabs}
\usepackage{amsmath}
\usepackage{cleveref}
\usepackage{multirow}
\usepackage{pifont}
\usepackage{tikz}
\usepackage{comment}
\usepackage{wrapfig}
\usepackage{enumitem}
\usepackage{subcaption}
\usepackage[outercaption]{sidecap}

\def\eg{\emph{e.g.}} 
\def\ie{\emph{i.e.}}

\newlength\savewidth\newcommand\shline{\noalign{\global\savewidth\arrayrulewidth
  \global\arrayrulewidth 1pt}\hline\noalign{\global\arrayrulewidth\savewidth}}

\usepackage{xpatch}
\makeatletter
\xapptocmd{\NAT@bibsetnum}{\setlength{\leftmargin}{0pt}\setlength{\itemindent}{\labelwidth}\addtolength{\itemindent}{\labelsep}}{}{}
\makeatother

\title{TIGeR: \underline{T}ext-\underline{I}nstructed \underline{Ge}neration and \underline{R}efinement \\ for Template-Free Hand-Object Interaction}

\author{
Yiyao Huang$^{1}$ \quad
Zhedong Zheng$^{2}$ \quad
Yu Ziwei$^{1}$ \quad
Yaxiong Wang$^{3}$ \quad \\
\textbf{Tze Ho Elden Tse}$^{1}$ \quad
\textbf{Angela Yao}$^{1}$ \quad \\
$^{1}$National University of Singapore, Singapore \\
$^{2}$University of Macau, Macau, China \\
$^{3}$Hefei University of Technology, Hefei, China \\
}

\begin{document}

\maketitle
\begin{tikzpicture}[remember picture,overlay]
\node[anchor=north west,xshift=3cm,yshift=-3.25cm] at (current page.north west) 
{\includegraphics[width=0.8cm]{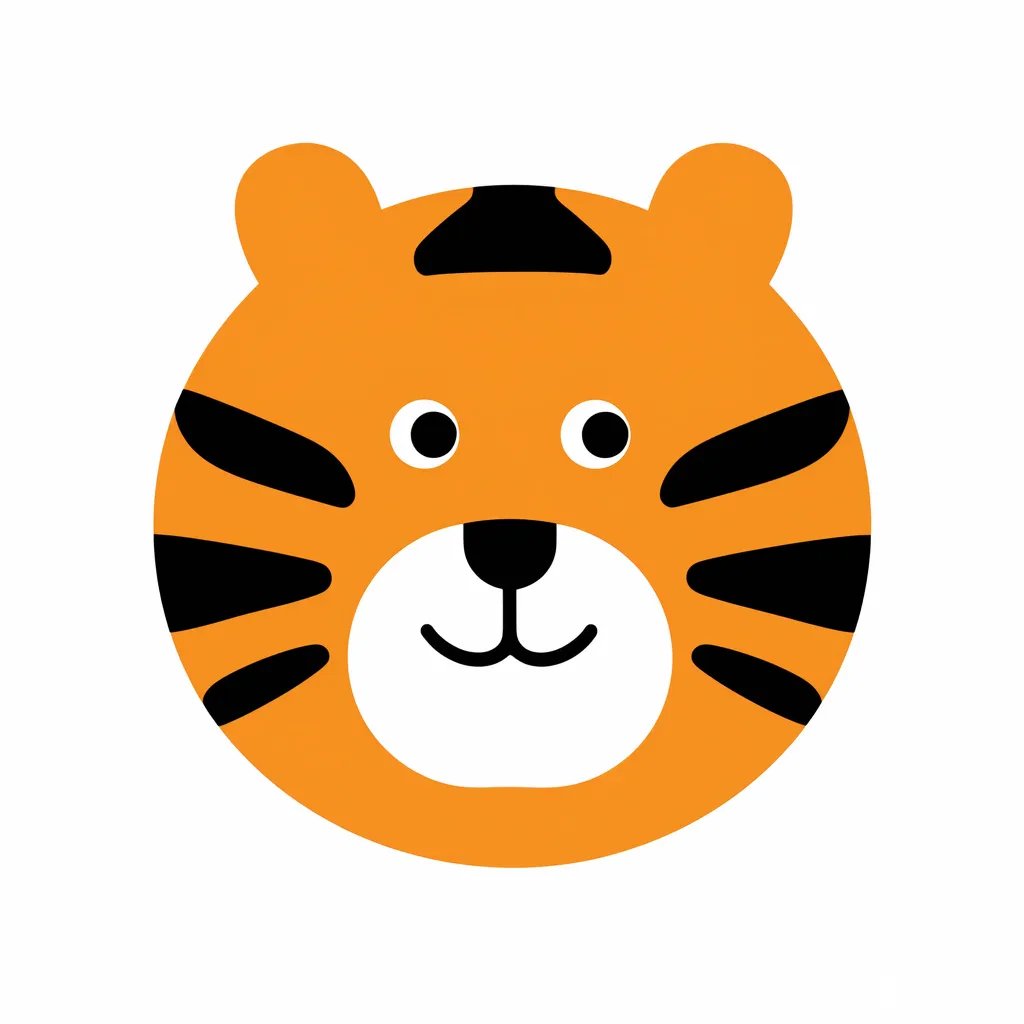}};
\end{tikzpicture}

\begin{abstract}
Pre-defined 3D object templates are widely used in 3D reconstruction of hand-object interactions.  However, they often require substantial manual efforts to capture or source, 
and inherently restrict the adaptability of models to unconstrained interaction scenarios, \eg, heavily-occluded objects. 
To overcome this bottleneck, we propose a new Text-Instructed Generation and Refinement (TIGeR) framework, harnessing the power of intuitive text-driven priors to steer the object shape refinement and pose estimation.  
We use a 
two-stage framework: a text-instructed prior generation and vision-guided refinement. 
As the name implies, we first 
leverage off-the-shelf models to generate shape priors  according to the text description without tedious 3D crafting.
Considering the geometric gap between the synthesized prototype and the real object interacted with the hand, we further calibrate the synthesized prototype via 2D-3D collaborative attention. 
TIGeR achieves competitive performance, \ie, \textbf{1.979}  and \textbf{5.468} object Chamfer distance on the widely-used Dex-YCB and Obman datasets, respectively, surpassing existing template-free methods. Notably, the proposed framework shows robustness to occlusion, while maintaining compatibility with heterogeneous prior sources, \eg, retrieved hand-crafted prototypes, in practical deployment scenarios.
\end{abstract}

\section{Introduction}

In this paper, we study 3D reconstruction of hand-object interactions in a monocular scene. Given a single-view RGB image containing interactive behavior, we predict the 3D point clouds of hands and objects.
This endeavor is crucial for enabling robots to comprehend and interact with the environment in a human-like manner, which serves as a key technology for applications such as Mobile ALOHA~\cite{Mobilealoha}. 
The undertaking necessitates a profound understanding of the input image and leverages the inherent 3D geometric structure priors of hands and objects to enhance the reconstruction quality. Since objects manipulated by hand have varied shapes, it is relatively challenging to obtain 3D prior knowledge of target object shapes. Some works, dubbed template-based methods~\cite{hasson2020leveraging, hasson2021towards}, directly apply predefined object templates, \eg, hand-crafted meshes, to hand-object interaction tasks. For instance, some researchers ~\cite{hasson2020leveraging} resort to ground-truth object templates from the YCB dataset~\cite{rhoi2020}, and only need to estimate 6D-pose of the given template to match input images. 
However, 3D object templates are usually inaccessible in real-world scenarios. 
Different from template-based methods, other studies~\cite{chen2022alignsdf,chen2021model,hasson19_obman,yu2022uv, tse2022collaborative}, referred to as template-free methods, recover UV maps or SDF representations from input RGB images. 
However, this line of methods typically suffers from self-occlusion by hand and thus fails to complete the entire object.
Inspired by the high-fidelity generation capabilities of cross-modal systems (particularly text-to-3D~\cite{yi2023progressive} and image-to-3D synthesis~\cite{Point-E,wang2024rigi}), we posit that a critical research question remains underexplored: Can synthesized 3D models function as viable foundational priors to encode generalized knowledge for open-world interaction scenarios?
As an early attempt to address this problem, we propose a Text-Instructed Generation and Refinement (TIGeR) framework that not only explores the prior generation pipeline but further bridges the gap between the generated prior and real-world observations.
In particular, our framework consists of 
two sequential stages: text-instructed prior generation and vision-guided refinement. 
Given a hand-object interaction image, we first apply a large multimodal question-answering (QA) model to obtain the description of the target object, and then leverage the cross-modal generative models to craft the corresponding shape prior. 
Next, we introduce a 2D-3D collaborative attention to fuse the 3D features of the shape prior and the 2D features of the input image. 
Based on the fused features, our model further refines point clouds to match the target object with geometric variants, if any. 
Finally, TIGeR involves the hand estimation, to co-optimize hand poses, hand meshes, and translations of both hand and objects.
Our method establishes correspondences between 3D point clouds and 2D images, enabling alignment for real hand-object interaction data. The entire process does not require any 3D template annotations, easing pre-requisites for real-world scenarios. 
Therefore, our contributions are as follows:

\begin{itemize}[leftmargin=0pt]
  \item \textbf{Template-free Framework.} Different from existing works demanding a pre-defined object template, we introduce a 
  Text-Instructed Generation and Refinement (TIGeR) framework to improve the scalability and ease the prerequisites for 3D hand-object interaction reconstruction. 
  Inspired by the recent success of text-based 3D object generation, we borrow the strength of text-driven prior to replace the hand-crafted template, and validate the feasibility.
  \item \textbf{Cookbook for Prior Refinement.} Given the gap between the 3D prior and the real object in the photo, 
  we introduce an attention-based paradigm to further register the object according to the visual cues.
  In particular, we integrate both 2D and 3D features via 2D-3D collaborative attention module,  
  simultaneously performing shape refinement and object registration.
  \item \textbf{Competitive and Robust Performance.} We evaluate our framework on two large-scale hand-object interaction datasets, \ie, Dex-YCB\cite{dexycb2021} and Obman~\cite{hasson19_obman}, surpassing other 
  competitive template-free approaches. Moreover, our method is robust against the common hand-occluded cases and is also scalable to other prototype sources, \eg, retrieved hand-crafted samples.
\end{itemize}

\begin{SCfigure*}[][t]
  {
  \caption[]{Here we show the input images, generated shape priors, predicted object and hand point clouds, and the corresponding 2D projections. 
  We could observe that the shape priors provide the common object geometry, which eases the further shape alignment. The proposed method, thus, achieves competitive reconstruction, especially for the heavy occlusions (\textbf{bottom}).
  } 
  \includegraphics[width=0.7\linewidth]{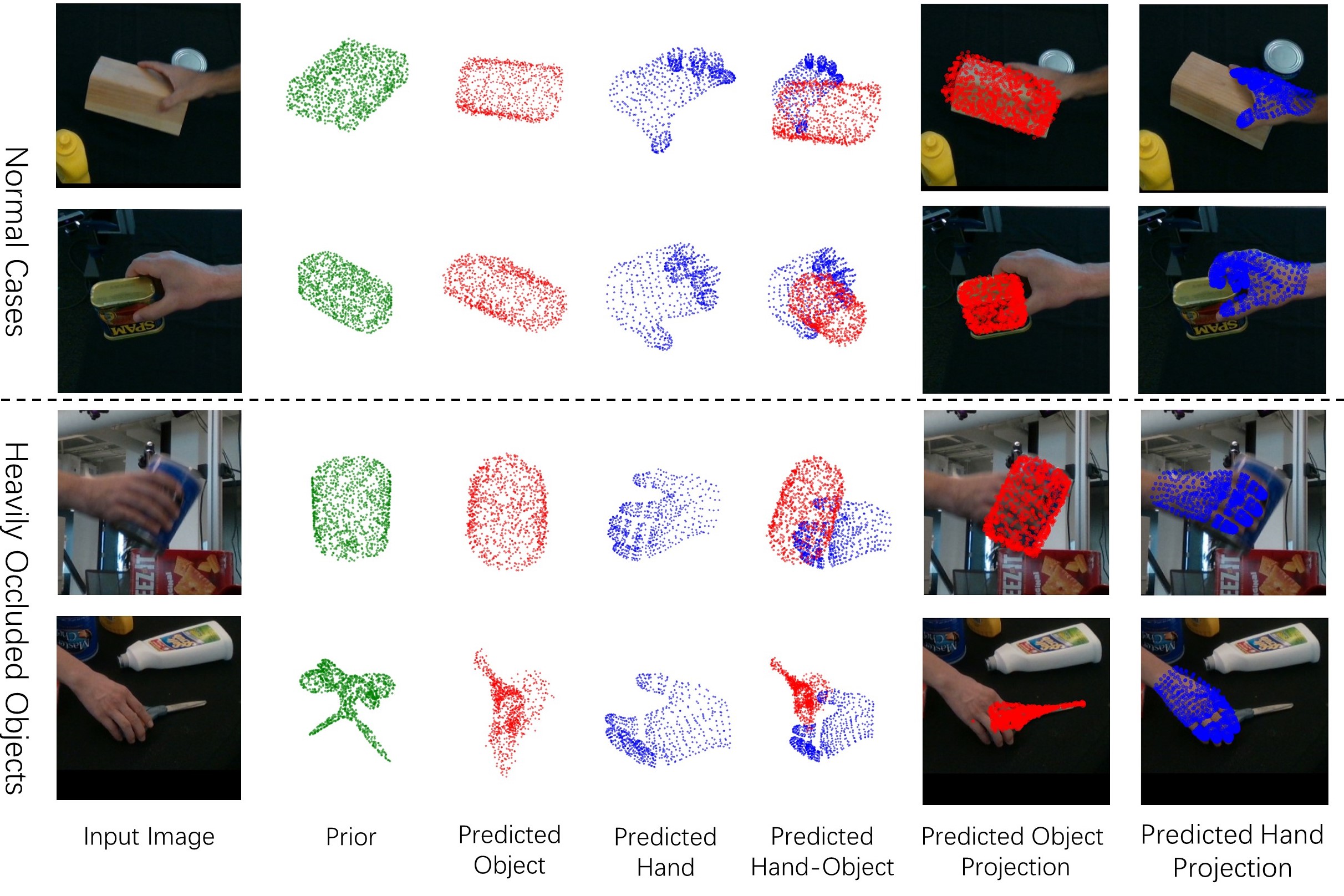}
  }
  \label{fig:one}
  \vspace{-.15in}
\end{SCfigure*}

\section{Related Work}

\noindent\textbf{3D hand pose and shape estimation.} 
Hand pose estimation methodologies have evolved through three technical paradigms. Early learning-based approaches~\cite{iqbal2018hand,mueller2018ganerated,tang2014latent,zimmermann2017learning} employ direct 3D keypoint regression from RGB inputs, and produce anatomically inconsistent surfaces that hinder downstream applications. 
This limitation motivates parametric modeling~\cite{smpl2015,mano2017}, \eg, MANO~\cite{mano2017} establishing a kinematic hand model. 
Besides, non-parametric paradigms~\cite{yang2019aligning,yang2019disentangling} circumvent shape space constraints through vertex-level prediction, employing disentangled autoencoders to isolate pose dynamics from background interference. Recent approaches integrate neural texture representations via UV mapping~\cite{Image-to-U2021,yu2022uv} and geometric attention mechanisms in transformer architectures~\cite{lin2021end,liu2022spatial,Li2022intaghand,tse2023spectral}, enabling joint optimization of skeletal pose and surface deformation.

\noindent\textbf{3D object reconstruction.} 
Early voxel-based approaches~\cite{Perspective_transformer_nets2016,deep_distangled_rep2017,marrnet2017} establish grid-form representations, yet remain constrained by cubic memory complexity. Subsequent approaches transitioned to point cloud representations~\cite{pointnet,pointnet++,point_transformer,pct2021}, employing graph-based aggregation modules to model local geometric structures. The field advances through implicit surface representations, with Park \textit{et al.}~\cite{deepsdf2019} pioneering memory-efficient shape encoding via continuous signed distance fields. Concurrent surface deformation strategies emerge, including FoldingNet's parameterized grid transformation~\cite{FoldingNet2018} and AtlasNet's MLP-driven mesh generation from primitive patches~\cite{AtlasNet2018}. Beyond geometric reconstruction, some works on pose estimation~\cite{densefusion,fsnet2021,posecnn2018,nocs2019} fuse RGB-D data to recover 6D object poses. 
Modern frameworks~\cite{corrI2P,deepI2P} instead perform cross-modal feature alignment, establishing geometric correspondences between 2D projections and 3D assets to derive pose parameters.

\noindent\textbf{3D hand-object interaction.} 
Recent works primarily fall into two categories: template-based and template-free approaches. Template-based methods~\cite{hasson2020leveraging,hflnet2023,hold2014} rely on RGB images paired with 3D object templates, leveraging multi-modal inputs for enhanced precision. Traditional pipelines~\cite{hold2014} employ hand pose regression followed by SfM initialization and refinement. Recent implementations extract global image features to estimate MANO parameters and 6D object poses~\cite{hasson2020leveraging}, while hybrid architectures combining single- and dual-stream backbones through ROIAlign operations~\cite{hflnet2023}.  Template-free approaches~\cite{hasson19_obman,chen2022alignsdf,chen2023gsdf} operate without explicit shape priors, directly predicting geometry from RGB inputs. Initial attempts deform a parametric sphere into target object surfaces using global image features~\cite{hasson19_obman,AtlasNet2018}, while contemporary methods integrate visual cues with pose information through signed distance field (SDF) decoders~\cite{chen2022alignsdf,chen2023gsdf}. Although these methods approximate real-world scenarios, their reconstructions remain susceptible to image degradation artifacts and occlusion-induced geometric ambiguities. Our framework addresses these limitations by incorporating text-guided shape priors from multimodal generative models, enabling detailed single-view reconstruction of manipulated objects via semantic-geometric alignment.


\section{Method}

\begin{figure}[t]
\centering
\vspace{-.15in}
\includegraphics[width=0.98\linewidth]{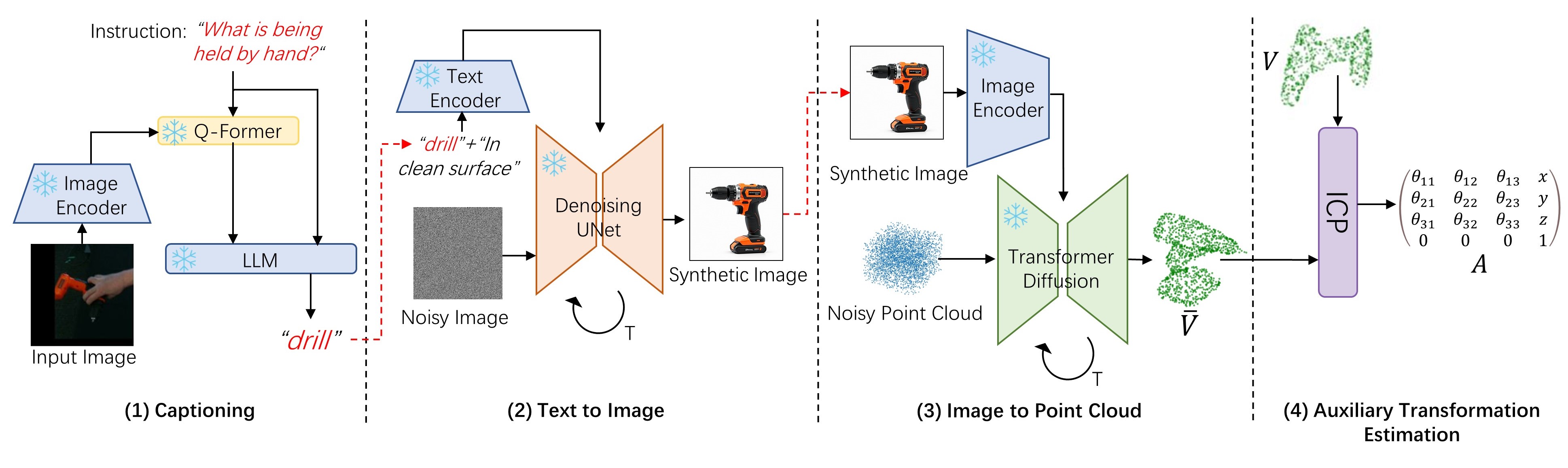}
\vspace{-.1in}
\caption[]{
A brief text-instructed prior generation pipeline. \textbf{(1) Captioning.} Given an input image depicting hand-object interaction, we first identify the occluded object by querying a multimodal large-scale model with the prompt: ``What is being held by hand?''
\textbf{(2) Text-to-Image Generation} Using the generated caption, we condition a diffusion model to synthesize a canonical view of the object without occlusions.
\textbf{(3) Image-to-Point Cloud Generation.} Finally, we employ an off-the-shelf 2D-to-3D lifting model to generate a 3D shape prior from the synthetic image.
\textbf{(4) Auxiliary Transformation Estimation}. We further estimate the auxiliary transformation $A$ between the shape prior and the ground-truth point cloud by Iterative Closet Point (ICP).
}
\addcontentsline{lof}{figure}{\numberline{\thefigure} The pipeline of acquiring the shape prior.}
\label{fig3}
\vspace{-.1in}
\end{figure}

Given a hand-object interaction image, our task is to reconstruct both 3D hand and object without relying on hand-crafted templates, which is usually inaccessible in real-world scenarios. 
Our framework contains two primary stages, \ie, text-instructed prior generation (see Fig.~\ref{fig3}), and vision-guided refinement (see Fig.~\ref{fig4}). During the first prior generation stage, we leverage off-the-shelf generative models to craft coarse shape prior $\bar{V}$ from the input image $I$. 
While this prior captures general semantic structure, it often lacks fine-grained geometric details aligned with the input image.
In the second vision-guided refinement stage, we intend to explicitly reduces the geometric discrepancy between the generated prior and the actual object in the image. 
In particular, we extract 2D visual features from the input image $I$ and 3D geometric features from the prior $\bar{V}$ and integrate both 2D and 3D features by leveraging 2D-3D collaborative attention modules. The fused features are then decoded into a refined point cloud $\tilde{V}$. Finally, we perform joint optimization of both the 3D hand and object to estimate their poses and output the final hand-object point clouds. In the following subsections, we elaborate two stages respectively.


\subsection{Text-instructed Prior Generation}\label{sec:prior}
\textbf{Prior Generation.} As shown in Figure~\ref{fig3}, our generation pipeline contains three phases: (1) Captioning, (2) Text-to-Image Generation, and (3) Image-to-Point Cloud Generation. 
We intend to obtain category information of target objects in the first phase. To this end, we query a pre-trained image caption model, \eg, InstrucBLIP~\cite{instrucBLIP}, using the prompt \textit{"What is being held by hand?"} The output text follows a specific format: \textit{"In this image, the hand is holding a [The category name of the object]."} Then, in the Text-to-Image Generation phase, we feed the structured text prompt \textit{"A [The category name of the object] in a clean surface"} into a text-to-image generator, \eg, Diffusion Model~\cite{flux2024}, to obtain a synthetic image with a clear background that only contains the target object. Next, we leverage the off-the-shelf image-to-point cloud model, \eg, Point-E~\cite{Point-E}, to generate the coarse-grained point cloud $\bar{V}$ as the 3D shape prior. Lastly, we apply Iterative Closest Point (ICP) to find the optimal transformation for $\bar{V}$.
\textbf{In this work, we do not pursue an optimal 3D prior but focus on validating the feasibility of the text-driven prior to replace the hand-crafted template.}

\textbf{Auxiliary Transformation Estimation.} To facilitate the model training, we also estimate the optimal transformation between the generated prior and the ground-truth mesh in the training set. In this way, we could have a pseudo one-to-one correlation during training to stabilize the model training in the early stage. Given the synthesized $\bar{V}$ and the ground-truth mesh $V$, we derive the optimal transformation matrix $A$ by Iterative Closest Point (ICP):
\begin{equation}
    A = \operatorname*{argmin}_A \lVert V - A\bar{V} \rVert_2^2.
\end{equation}
Given the predicted transformation $A$,  we could have a pseudo one-to-one mapping between synthesized $\bar{V}$ and the ground-truth mesh $V$ as $\mathcal{J}(i) = \operatorname*{argmin}_{j} \lVert V_i - A\bar{V}_j \rVert_2^2 $.  $\mathcal{J}(i)$ denotes the index of $\bar{V}_j$, which is the nearest neighbor of $V_i$. \textbf{We note that we do not use such estimation during inference. The auxiliary pseudo transformation is only estimated for training.}

\subsection{Vision-guided Refinement}\label{sec:registration}

\begin{figure}[t]
\vspace{-.15in}
\centering
\includegraphics[width=0.97\linewidth]{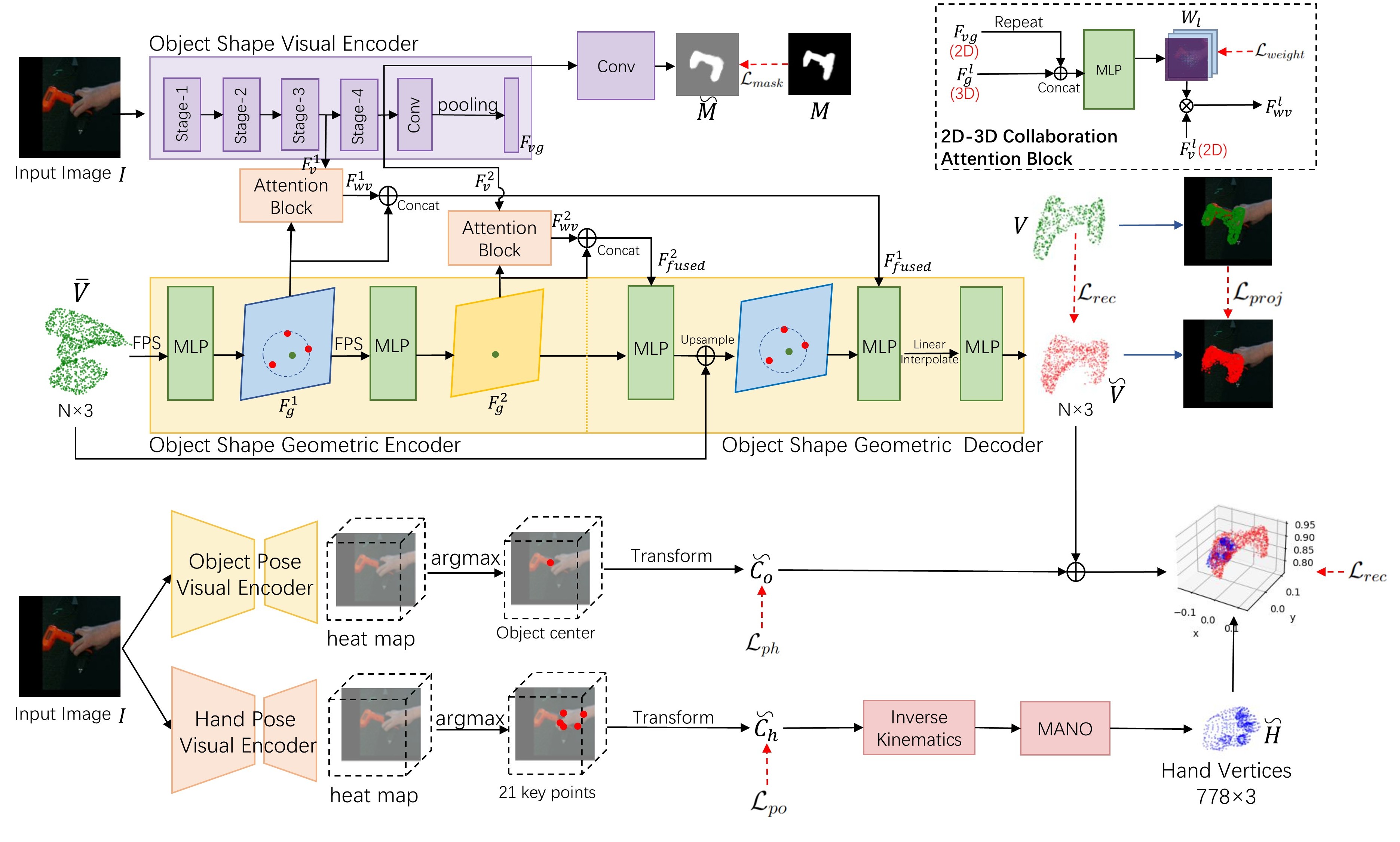}
\vspace{-.1in}
\caption[]{
Overview of vision-guided refinement stage. \textbf{Top:} Given the text-driven shape prior $\bar{V}$ and the RGB image $I$, we extract the 2D visual feature via object shape visual encoder and the 3D geometric feature via object shape geometric encoder. 
Then we apply 2D-3D collaboration attention blocks (\emph{top right}) to fuse the visual feature and the geometric feature. The fused features are then fed to the object shape geometric decoder to predict the object shape $\tilde{V}$. 
\textbf{Bottom:} Given the input image $I$, we estimate the object center in the image and hand poses, \ie, 21 key points. On one hand, the input image is fed to the object pose visual encoder, which does not share weight with the object shape visual encoder, to obtain the center estimation. On the other hand, we apply the hand pose visual encoder to predict 21 key points. We then manipulate the MANO model to reconstruct the hand $\tilde{H}$. Finally, we fuse the object point cloud and hand point cloud supervised by $\mathcal{L}_{rec}$.
}
\addcontentsline{lof}{figure}{\numberline{\thefigure} Overview of shape refinement stage.}
\label{fig4}
\vspace{-.1in}
\end{figure}

\noindent\textbf{Shape refinement.} As shown in Figure~\ref{fig4}, we show the brief structure of our vision-guided refinement stage. 
Given a coarse shape prior $\bar{V}$ and an input image $I$, we first extract complementary 2D and 3D features through dedicated visual and geometric encoders. 
Since shape prior $\bar{V}$ contains category-level geometric knowledge, such as the cuboid structure of boxes, the object shape geometric encoder processes $\bar{V}$ through two hierarchical layers, producing local features $F_g^1$ and $F_g^2$. 
Similarly, we obtain multi-resolution visual features from the input image $I$ via the object shape visual encoder, yielding local visual features $F_v^1$, $F_v^2$ and global feature $F_{vg}$ via average pooling. 
To align the shape prior $\bar{V}$ with the hand-object interaction scene, we propose a cross-modal feature fusion approach that establishes correspondences between 3D patches and 2D image regions. The fusion process begins by repeating and concatenating the global visual feature $F_{vg}$ with each 3D patch's geometric features to form an initial fused representation. The fused representation is then processed by MLPs followed by softmax to generate attention weights $W_l, l\in\{1,2\}$, which identify the relevant image regions for each 3D patch. $W_l$ are applied to the visual features $F_{wv}^l$. We finally concatenate $F_{wv}^l$ with $F_{g}^l$ to get the fused feature $F_{fused}^l$, which contains both precise information from 3D patches and rich visual cues from the corresponding image regions. Given the fused local feature $F_{fused}^l$, the object shape geometric decoder predicts adjusted 3D coordinates to align the shape prior with the interaction scene. The decoding, following U-net~\cite{unet} style, consists of two processes. First, $F_{fused}^2$ is processed through MLPs and interpolated to generate features for intermediate points. The $F_{fused}^1$ is then concatenated with these intermediate features and passed through additional MLPs, followed by linear interpolation to complete features for all remaining points. Finally, the network regresses 3D coordinates for every vertex to produce the aligned object point cloud $\tilde{V}$.

\noindent\textbf{Pose estimation.} Simultaneously, we predict the object center location and the hand pose through two independent object and hand pose visual encoders, respectively.
As shown in the bottom of Figure~\ref{fig4}, we apply the hand pose visual encoder to extract 21 feature maps from the input image $I$. 
Then, we obtain the uvd (u+v+depth) location of the max activation values in every heatmap as 21 hand key points. 
Given the camera intrinsic, the uvd coordinates can be transformed into 3D positions $\tilde{C_h}$ in the world coordinate system. 
We apply Inverse kinematics~\cite{inverse_kinematics} to convert $\tilde{C_h}$ into MANO parameters, which are then provided to the MANO model to get the hand vertices $\tilde{H}$.  Similarly, given the input image, we apply the object pose visual encoder to extract the object heat map, and then obtain the index for the point with the maximum activation value. 
Then we transform the point index into the 3D coordinates of the object center $\tilde{C_o}$.
We translate the refined object $\tilde{V}$ to the predicted center, and compose the reconstructed object and hand as the final output.

\textbf{Optimization objectives.}
To facilitate reconstructing the geometric shape of the target object in the early training, we introduce several auxiliary tasks. For instance, we leverage the pseudo one-to-one mapping $\mathcal{J}(i)$ (defined in Section~\ref{sec:prior}) from the point index of the target object $V$ to the point index of the shape prior $\bar{V}$ to supervise the intermediate attention weight $W_{2}$, which is in the second 2D-3D Collaboration Attention Block as:
\begin{equation}
    \mathcal{L}_{weight} = \frac{1}{|V|} \sum_{i, j = \mathcal{J}(i)} \|\phi(V_i) - argmax(W_2{(j)}))||^2_2,
    \label{attention loss}
\end{equation}
where $\phi$ denotes the 2D projection of the vertex in the ground-truth $V$. The second term is the coordinates of max activation in the corresponding heatmap $W_2$. 
Similarly, we apply the pseudo one-to-one mapping to supervise the projection of the final reconstructed object as:
\begin{equation}
    \mathcal{L}_{proj} = \frac{1}{|V|}\sum_{i, j = \mathcal{J}(i)} \|\phi(V_i) - \phi(\tilde{V}_j)\|,
\end{equation}
For 3D supervision, we apply the conventional group-to-group reconstruction loss via Chamfer Distance, which can be derived as :
\begin{equation}
    \mathcal{L}_{rec} = \frac{1}{|\tilde{V}|} \sum_{i=1}^{|\tilde{V}|} \min_{j} \| \tilde{V}_i - V_j \|_2^2 + \frac{1}{|V|} \sum_{j=1}^{|V|} \min_{i} \| V_j - \tilde{V}_i \|_2^2
\end{equation}

Furthermore, we introduce foreground mask supervision as an auxiliary task to make our object shape visual encoder concentrate on the target object and mitigate the negative impact of occlusion. 
Specifically, we take $F_v^2$ as input followed by a 2D-convolutional layer, a max pooling layer and sigmoid function to estimate $\tilde{M}$ as the foreground probability. 
The foreground mask loss is a binary classification task, which can be formulated as:
\begin{equation}
    \mathcal{L}_{mask} = -\sum_i (M_ilog(\tilde{M}_i)+(1-M_i)log(1-\tilde{M}_i)), 
\end{equation}
where $M$ is the resized ground-truth amodal mask. For the two pose visual encoders, we introduce $\mathcal{L}_{ph}$ and $\mathcal{L}_{po}$ as L2 distance between the prediction $\tilde{C}$ and the corresponding ground truth $C$ as: 
\begin{equation}
    \mathcal{L}_{ph} = ||C_h - \tilde{C_h}||^2_2,  \mathcal{L}_{po}= ||C_o - \tilde{C_o}||^2_2.
\end{equation}
Therefore, the final loss function for the shape refinement stage is:
\begin{equation}
    \mathcal{L}_{registration} = \mathcal{L}_{rec} + \mathcal{L}_{mask} + \mathcal{L}_{ph} + \mathcal{L}_{po} + \lambda_{weight}\mathcal{L}_{weight} + \lambda_{proj}\mathcal{L}_{proj},
\end{equation}
Considering that $\mathcal{L}_{weight}$ and $\mathcal{L}_{proj}$ are based on the pseudo alignment, we empirically set a relatively small weight,   \ie, $\lambda_{weight} = 0.1$, $\lambda_{proj} = 0.01$.

\section{Experiment}
\textbf{Implementation details.} The whole hand-object reconstruction process contains two stages. (1) In the text-instructed prior generation stage, we leverage three pre-trained off-the-shelf generative models to craft shape priors for the given RGB images. We adopt InstructBLIP~\cite{instrucBLIP} for captioning, FLUX-1~\cite{flux2024} for image synthesis, and Point-E~\cite{Point-E} for point cloud generation. \textbf{In this work, we do not pursue the optimal prior, but validate the effectiveness of the generated priors. The proposed method is compatible with different prior sources (see Section~\ref{sec:ablation}).} (2) In the vision-guided refinement stage, we harness pre-trained HRNet~\cite{hrnet2021} and Pointnet++~\cite{pointnet++} as visual and geometric backbones to extract the shape features of objects. We adopt Resnet50~\cite{resnet} as the backbone for both object and hand pose visual encoder.  
We train our model for 1,600 epochs using Adam~\cite{Kingma2014AdamAM} with the initial learning rate of $5e^{-5}$  on 4 Nvidia RTX A5000 GPUs. 
To eliminate the reconstruction error after assembling the object and hand, we further fine-tune the entire framework for another 100 epochs using $\mathcal{L}_{rec}$ with the learning rate of $1e^{-5}$, while freezing the object shape visual encoder and the hand pose visual encoder.



\subsection{Comparison with the State-of-the-Art Methods}

\begin{figure}[]
\centering
\vspace{-.15in}
\includegraphics[width=0.95\linewidth]{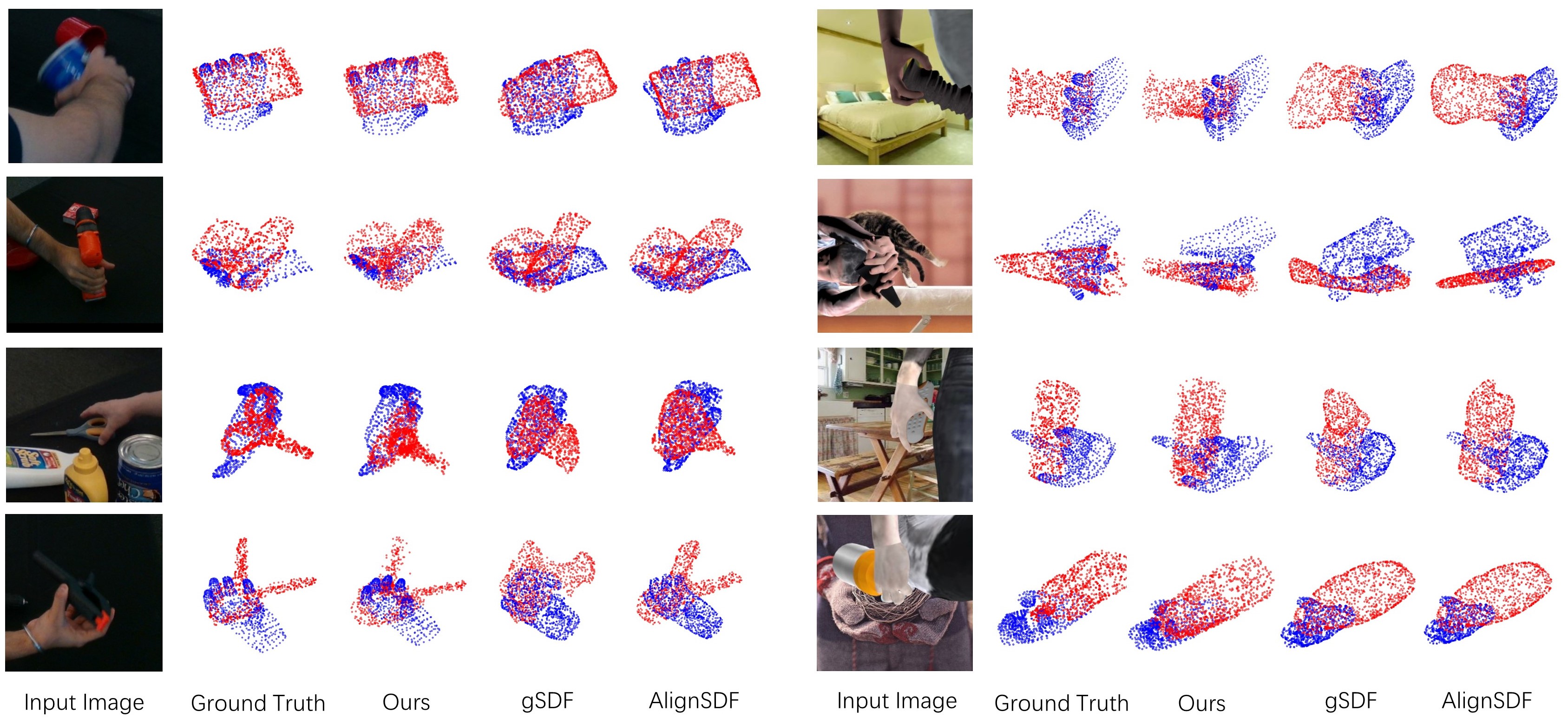}
\vspace{-.1in}
\caption{
Qualitative comparision of TIGeR (Ours) and prevailing template-free methods, including gSDF~\cite{chen2023gsdf}, AlignSDF~\cite{chen2022alignsdf}, and Hasson~\cite{hasson19_obman} on DexYCB (\emph{left}) and Obman (\emph{right}).
}
\label{fig6_1}
\vspace{-.15in}
\end{figure}

As shown in Table~\ref{table:one} and Table~\ref{table:two}, we could observe that the quality of objects reconstructed by our method surpasses the quality of objects produced by template-free SOTA methods on both the DexYCB dataset and the Obman dataset. 
For instance, our method has arrived at 1.979 median Chamfer Distance ($CD_o$) 0.292 $FS_o$@5 and 0.637 $FS_o$@10 on the DexYCB dataset, which surpasses gSDF~\cite{chen2023gsdf} by a clear margin. 
We observe a similar phenomenon on the Obman dataset. Our method has achieved  5.468  $CD_o$, and competitive $0.199$ $FS_o$@5 and $0.462$ $FS_o$@10 scores.
As for reconstructed hands, on both datasets, our method yields high-quality hands with the lowest median Chamfer Distance ($CD_h$), while yielding the highest $FS_h@1$ and $FS_h@5$.
Furthermore, we show the qualitative comparison of our methods and SOTA methods in Figure~\ref{fig6_1}. 
Our method achieves superior geometric fidelity for both simple primitives (\eg, cans, boxes) by preserving sharp edges and planar surfaces, and complex articulated objects (\eg, scissors, drills) through high-fidelity detail retention, whereas competitive methods exhibit significant shape distortions and topological oversimplification.
It is worth noting that SDF-based methods generate uniformly distributed points in the reconstructed surface geometry, resulting in geometrically ambiguous reconstructions of articulated hand regions. This inherent uniformity inadequately captures the non-linear deformation patterns required for dexterous finger manipulation in real-world scenarios.  In contrast, the proposed method leverages the straightforward kinematic-aware hand parametric model and generated object priors to ease the optimization difficulty, while preserving more interaction details. 

\begin{table*}[t]
\centering
\small
\begin{tabular}{c|cccccc}
\shline
Method & $CD_o\downarrow$ & $FS_o@5\uparrow$ & $FS_o@10\uparrow$
       & $CD_h\downarrow$ & $FS_h@1\uparrow$ & $FS_h@5\uparrow$\\
\hline
Hasson~\cite{hasson19_obman} & 5.831 & 0.155 & 0.405 & 6.375 & 0.003 & 0.162 \\
AlignSDF~\cite{chen2022alignsdf} & 2.669 & 0.254 & 0.588 & 2.768 & 0.003 & 0.222 \\
gSDF~\cite{chen2023gsdf}     & 2.769 & 0.258 & 0.591 & 2.770 & 0.003 & 0.222 \\
\hline
TIGeR (Ours)     & \textbf{1.979} & \textbf{0.292}  & \textbf{0.637} & \textbf{1.132} & \textbf{0.008} & \textbf{0.413}  \\
\shline
\end{tabular}
\vspace{-.1in}
\caption{Quantitative results of hand-object reconstruction performance on DexYCB.
} \label{table:one}

\centering
\small
\begin{tabular}{c|cccccc}
\shline
Method & $CD_o\downarrow$ & $FS_o@5\uparrow$ & $FS_o@10\uparrow$
       & $CD_h\downarrow$ & $FS_h@1\uparrow$ & $FS_h@5\uparrow$\\
\hline
Hasson19~\cite{hasson19_obman}$^*$ & - & - & - & - & - & - \\
AlignSDF~\cite{chen2022alignsdf} & 5.584 & 0.203 & 0.476 & 2.117 & 0.004 & 0.248 \\
gSDF~\cite{chen2023gsdf} & 5.626 & \textbf{0.207} & \textbf{0.482} & 2.116 & 0.004 & 0.249 \\
\hline
TIGeR (Ours) & \textbf{5.468} & 0.199 & 0.462 & \textbf{0.787} & \textbf{0.013} & \textbf{0.537} \\
\shline
\end{tabular}
\vspace{-.1in}
\caption{Quantitative results of hand-object reconstruction performance on Obman. $^*$: We re-implement the official code but the method does not converge when involving both hand and object.
} \label{table:two}
\end{table*}

\begin{table*}[t]
\vspace{-.15in}
\centering
\small
\begin{tabular}{c|ccc|ccc}
\shline  
\multirow{2}{*}{Method} & \multicolumn{3}{c}{DexYCB} & \multicolumn{3}{|c}{Obman}\\
 & $CD_o\downarrow$ &$ FS_o@5\uparrow$ & $FS_o@10\uparrow$ & $CD_o\downarrow$ &$ FS_o@5\uparrow$ & $FS_o@10\uparrow$\\ 
\hline  
Hasson~\cite{hasson19_obman}     & 1.17 & 0.36 & 0.78 & 2.90 & 0.27 & 0.61 \\
AlignSDF~\cite{chen2022alignsdf} & 1.41 & 0.37 & 0.73 & 3.65 & 0.24 & 0.54 \\
gSDF~\cite{chen2023gsdf}         & 1.53 & 0.36 & 0.72 & 3.88 & 0.23 & 0.53 \\
\hline 
TIGeR (Ours) & \textbf{0.62} & \textbf{0.54} & \textbf{0.90} & \textbf{2.78} & \textbf{0.30} & \textbf{0.63} \\
\shline
\end{tabular}
\vspace{-.1in}
\caption{Comparison of reconstructed object \textbf{only}. We have centerlize all objects for a fair comparison. 
} \label{table:four}
\centering
\end{table*}

\subsection{Ablation Studies and Further Discussion}\label{sec:ablation}
\begin{wrapfigure}{r}{0.48\textwidth}
\vspace{-.1in}
  \centering
  \includegraphics[width=0.45\textwidth]{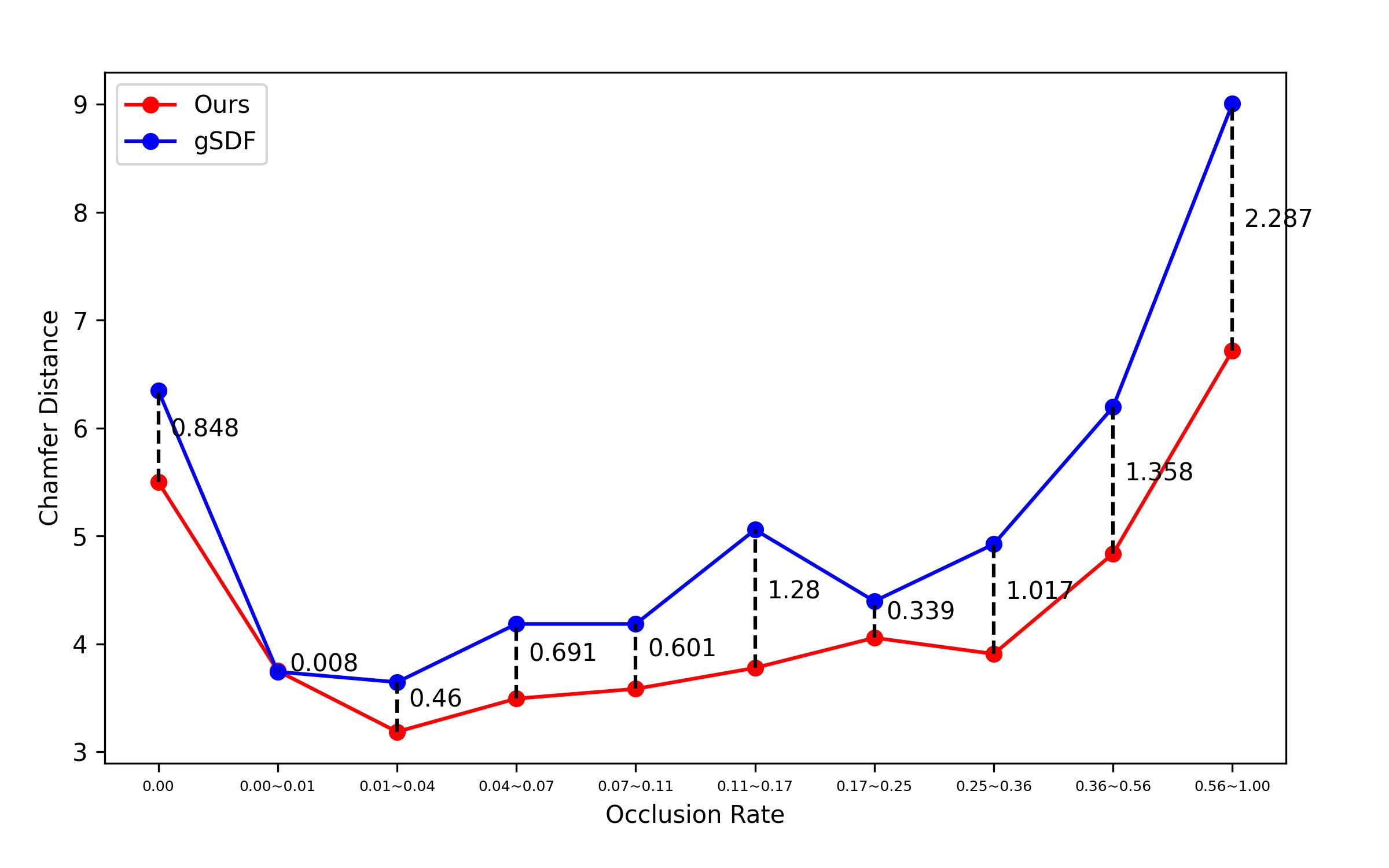}
  \vspace{-.15in}
  \caption{Comparison between ours and the competitive gSDF against the occlusion. 
  We could observe that \textcolor{red}{our method} has achieved lower Chamfer distance than \textcolor{blue}{gSDF} in all ranges of occlusion rate, especially for heavy occlusion. 
  }
  \label{fig9}
  \vspace{-.1in}
\end{wrapfigure}

\textbf{Comparison of the object reconstruction only.} To isolate object reconstruction quality, we center-normalize both predicted and ground-truth objects by aligning their centroids to the origin. Our re-implementation of three competitive methods reveals that Hasson~\cite{hasson19_obman} achieves optimal $\text{CD}_o$ (1.17/2.90) and $\text{FS}_o$ scores when evaluated purely on object reconstruction. As shown in Table~\ref{table:four}, our method further reduces the Chamfer distance as $0.62$ on DexYCB and $2.78$ on Obman, surpassing Hasson by a clear margin.

\begin{wrapfigure}{r}{0.48\textwidth}
\vspace{-.1in}
  \centering
  \includegraphics[width=0.45\textwidth]{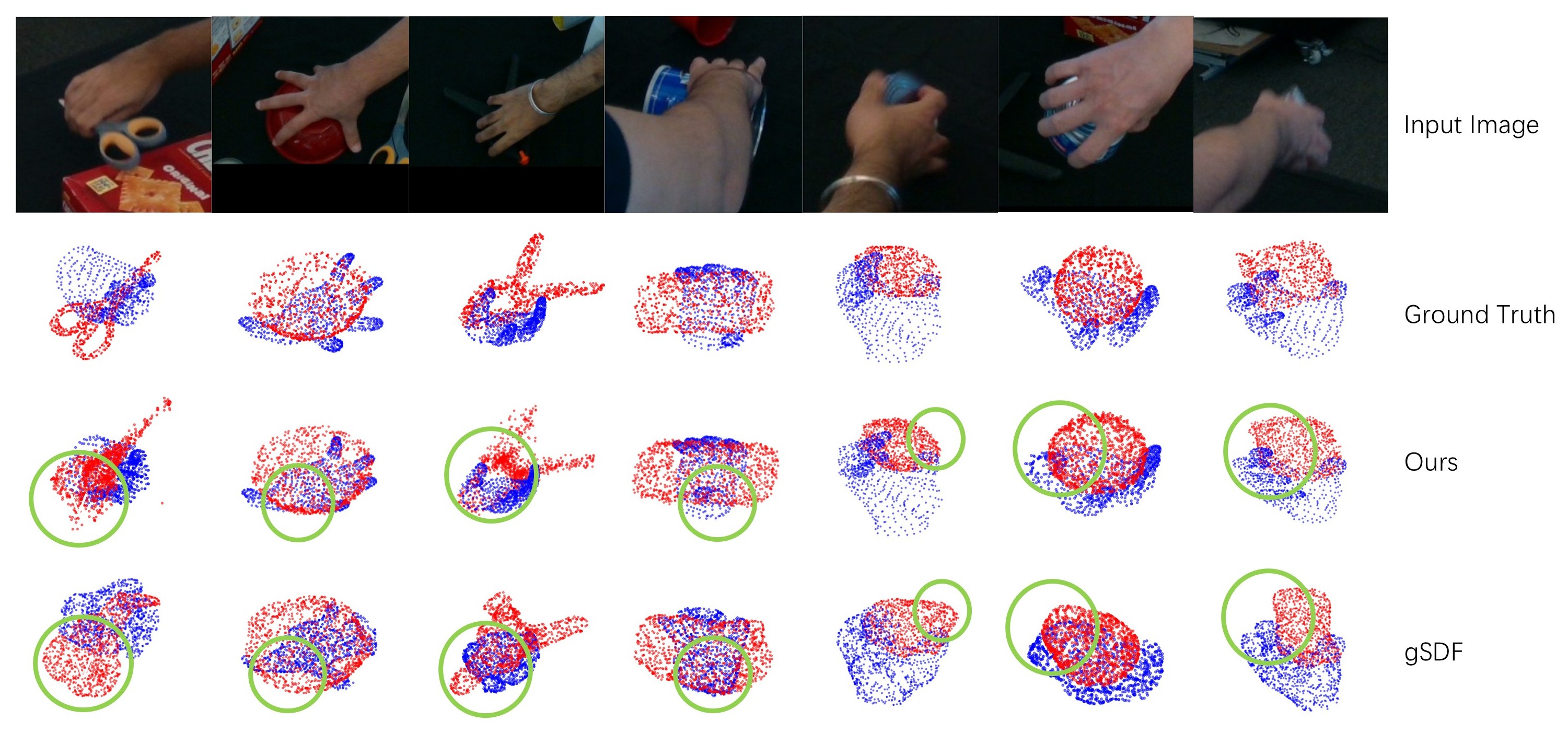}
  \caption{
    Qualitative comparison between our method and gSDF under severe occlusion scenarios. The green circles highlight the prediction discrepancy. 
    }
  \label{fig13}
  \vspace{-.15in}
\end{wrapfigure}

\textbf{Robustness against occlusions.} The target objects interacted by the human are usually occluded by hands or other objects. 
We analyze the relationship between the reconstruction quality of objects and the degree of occlusion. We employ the ground truth amodal mask $M_{amodal}$ and visible mask $M_{visible}$ of the target objects to measure the occlusion rate of testing samples as $ R_{occlusion} = 1 - \frac{Area(M_{visible}) + 1}{Area(M_{amodal})+1},$ where Area($\cdot$) denotes the area of the foreground in the  corresponding mask. 
We split the samples into 10 equal-numbered groups according to their occlusion rate. 
In Figure~\ref{fig9}, we 
report the median Chamfer Distance between the predicted point clouds and the ground truth point clouds for every group. 
As the increasing occlusion rate, the proposed method yields a clear margin towards competitive gSDF. As shown in Figure~\ref{fig13}, we visualize some samples with severe occlusion compared to gSDF. 
This robustness stems from our shape prior to provide geometric cues: (1) For simple geometric objects, \eg, cans and boxes, the prior effectively preserves sharp edges and planar surfaces even under heavy occlusion; (2) For complex articulated objects, \eg, scissors, the prior maintains proper handle and blade geometry. We highlight the discrepancy in Figure~\ref{fig13} with green circles.


\begin{figure}[]
\centering
\vspace{-.15in}
\includegraphics[width=0.9\linewidth]{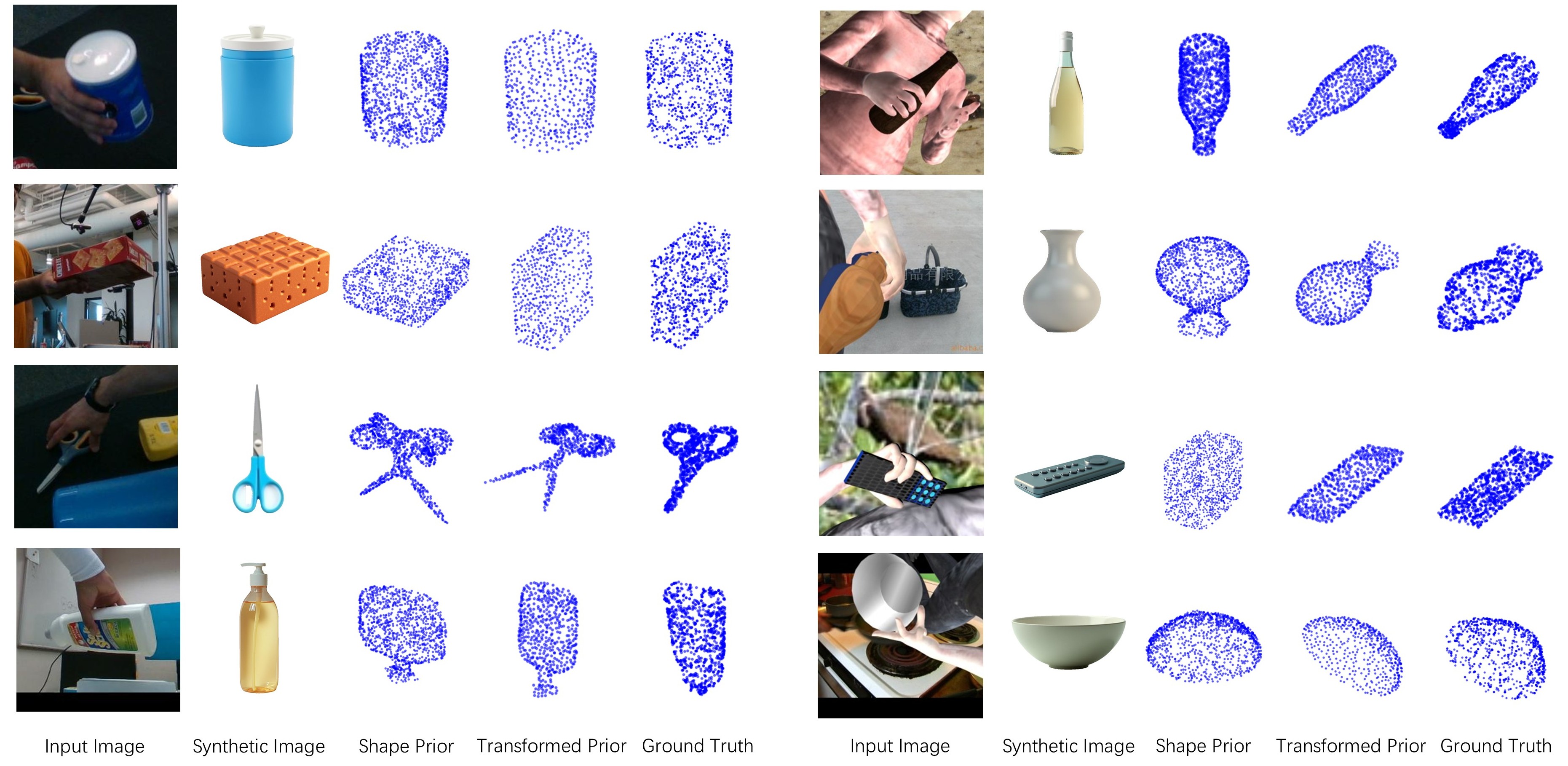}
\caption{
Here we show the intermediate during generation, including the text-to-image result (\ie, synthetic image), image-to-3D result (\ie, shape piror), and the pseudo transformation (\ie, transformed priors) and ground-truth object on the training set of DexYCB (\emph{left}) and Obman (\emph{right}). 
}
\vspace{-.15in}
\label{fig5}
\end{figure}

\begin{figure}[t]
  \centering
  \vspace{-.15in}
  \begin{subfigure}[t]{0.45\textwidth}
    \centering
    \includegraphics[width=\linewidth]{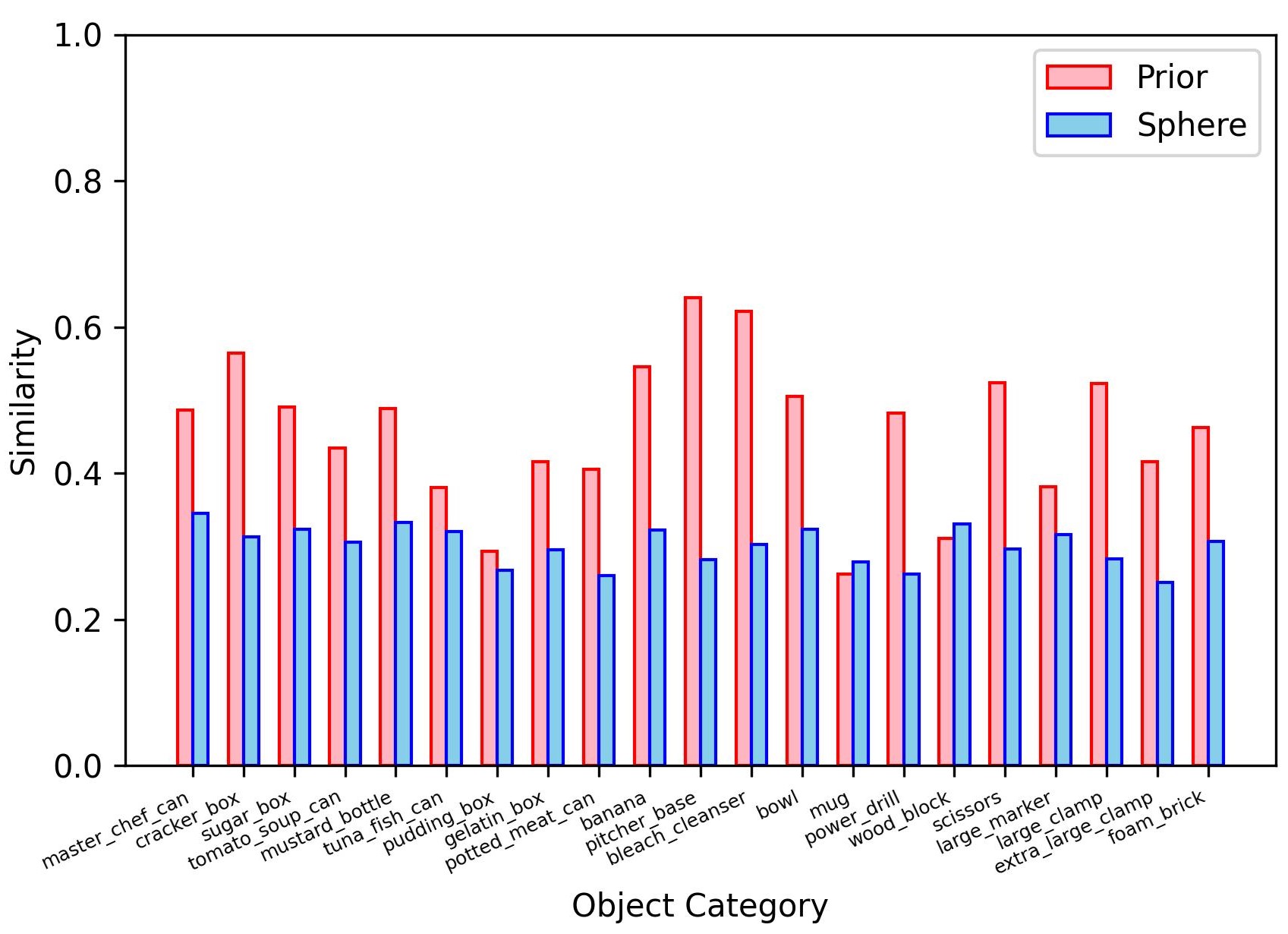} 
    \vspace{-.2in}
    \caption{Without Transformation}
    \label{fig10_1}
  \end{subfigure}
  \hfill
  \begin{subfigure}[t]{0.45\textwidth}
    \centering
    \includegraphics[width=\linewidth]{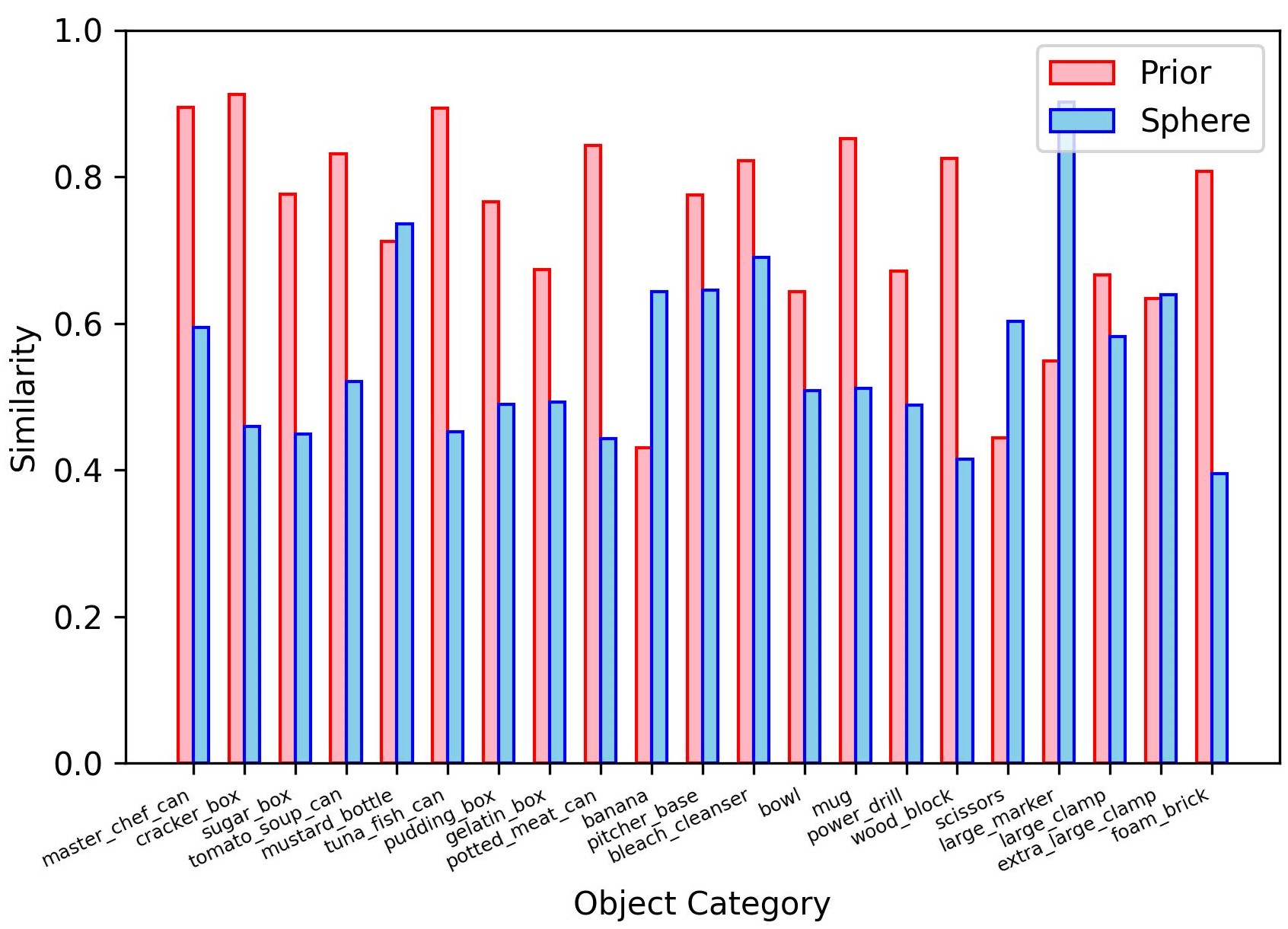}
    \vspace{-.2in}
    \caption{With Pseudo Transformation}
    \label{fig10_2}
  \end{subfigure}
  \vspace{-.1in}
  \caption{Similarity between our prior and ground-truth template (\textcolor{red}{red}), sphere and ground-truth template (\textcolor{blue}{blue}) in terms of different object categories on the training set of the DexYCB dataset. \textbf{Higher is better.} We observe that whether the transform is performed or not, the generated prior is more similar to the ground truth than the widely-used sphere initialization, facilitating the optimization. }
  \label{fig:mainfigure}
  \vspace{-.1in}
\end{figure}

\noindent\textbf{Study of prior quality.}\label{sec4.1.2} We show intermediate results of the first text-instructed prior generation stage in Figure~\ref{fig5}. We observe that our generation result gradually approaches the target object. We also quantitatively study the generated prior quality by comparing it with the commonly-used unit sphere.  In particular, we adopt  DGCNN \cite{dgcnn2019}  to extract perceptual features, and calculate the feature similarity with the ground-truth. 
As shown in Figure~\ref{fig10_1}, we find that the generated prior easily surpasses the sphere unit. We further apply the pseudo transformation to both our prior and the sphere in Figure~\ref{fig10_2}. The proposed method yields a higher  similarity score among all subcategories.



\setlength{\tabcolsep}{3pt}
\begin{table}[t]
\centering
\caption{\textbf{Ablation studies.} \textbf{(a)} We adopt different prior sources for comparison. \textbf{(b)} We study the effect of two projection losses. \textbf{(c)} We study the performance of four different sub-categories based on the proposed prior and commonly-used sphere prototype. }
\label{table:main}
\vspace{-.15in}
\begin{minipage}{0.4\textwidth} 
    \begin{subtable}[t]{0.9\textwidth}
        \centering
        \small
        \caption{}
        \vspace{-.1in}
        \begin{tabular}{c|ccc}
        \shline  
        Source of prior & $CD_o\downarrow$ &$ FS_o@5\uparrow$ & $FS_o@10\uparrow$ \\
        \hline  
        Unit Spheres & 0.67 & 0.53 & 0.88 \\
        Retrieval Images & 0.65 & 0.54 & 0.90 \\
        Synthetic Images & \textbf{0.62} & \textbf{0.55} & \textbf{0.90} \\
        \shline
        \end{tabular}
        \label{table:five}
    \end{subtable}

    \begin{subtable}[t]{0.9\textwidth}
        \centering
        \small
        \caption{}
        \vspace{-.1in}
        \begin{tabular}{cc|cccc}
        \shline  
        $\textbf{L}_{weight}$ & $\textbf{L}_{proj}$ & $CD_o\downarrow$ &$ FS_o@5\uparrow$ & $FS_o@10\uparrow$ \\
        \hline  
         \ding{55} & \ding{51} & 3.88 & 0.24 & 0.57 \\
         \ding{51} & \ding{55} & 1.32 & 0.38 & 0.77 \\
         \ding{51} & \ding{51} & \textbf{0.62} & \textbf{0.55} & \textbf{0.90} \\
        \shline
        \end{tabular}
        \label{table:seven}
    \end{subtable}
\end{minipage}
\hfill
\begin{minipage}{0.4\textwidth} 
    \begin{subtable}[t]{0.9\textwidth}
        \centering
        \small
        \caption{}
        \vspace{-.1in}
        \begin{tabular}{c|cc|cc}
        \shline  
        \multirow{2}{*}{Category} & \multicolumn{2}{c|}{Prototype} & \multirow{2}{*}{$Sim.\uparrow$} & \multirow{2}{*}{$CD_o\downarrow$} \\
                 & Prior & Sphere & & \\
        \hline  
        \multirow{2}{*}{Boxes} & \ding{51} & & \textbf{0.775} & \textbf{0.585} \\
                               & & \ding{51} & 0.496 & 0.635 \\
        \hline
        \multirow{2}{*}{Cans}  & \ding{51} & & \textbf{0.707} & 0.585 \\
                               & & \ding{51} & 0.563 & \textbf{0.572} \\
        \hline
        \multirow{2}{*}{Bottles}   & \ding{51} & & \textbf{0.743} & \textbf{0.639} \\
                                   & & \ding{51} & 0.639 & 0.705 \\
        \hline
        \multirow{2}{*}{Others}    & \ding{51} & & \textbf{0.712} & \textbf{0.678} \\
                                   & & \ding{51} & 0.561 & 0.780 \\
        \shline
        \end{tabular}
        \label{table:six}
    \end{subtable}
\end{minipage}
\vspace{-.15in}
\end{table}

 \textbf{
 Scalability to different prior sources.} 
 To validate that our framework is compatible with different prior sources, we adopt the retrieved images to replace the synthetic images during the prior generation. As shown in Table~\ref{table:five}, we observe that priors from synthetic images also perform well, surpassing the baseline with unit sphere by a clear margin. 
 We further analyze the performance of different object categories (\ie, boxes, cans, bottles, others on DexYCB) in Table~\ref{table:six}. 
 Except for the sphere-alike `can' objects, our prior usually achieves lower median Chamfer Distance than the unit sphere.

\begin{wrapfigure}{r}{0.48\textwidth}
  \centering
    \vspace{-.15in}
  \includegraphics[width=0.45\textwidth]{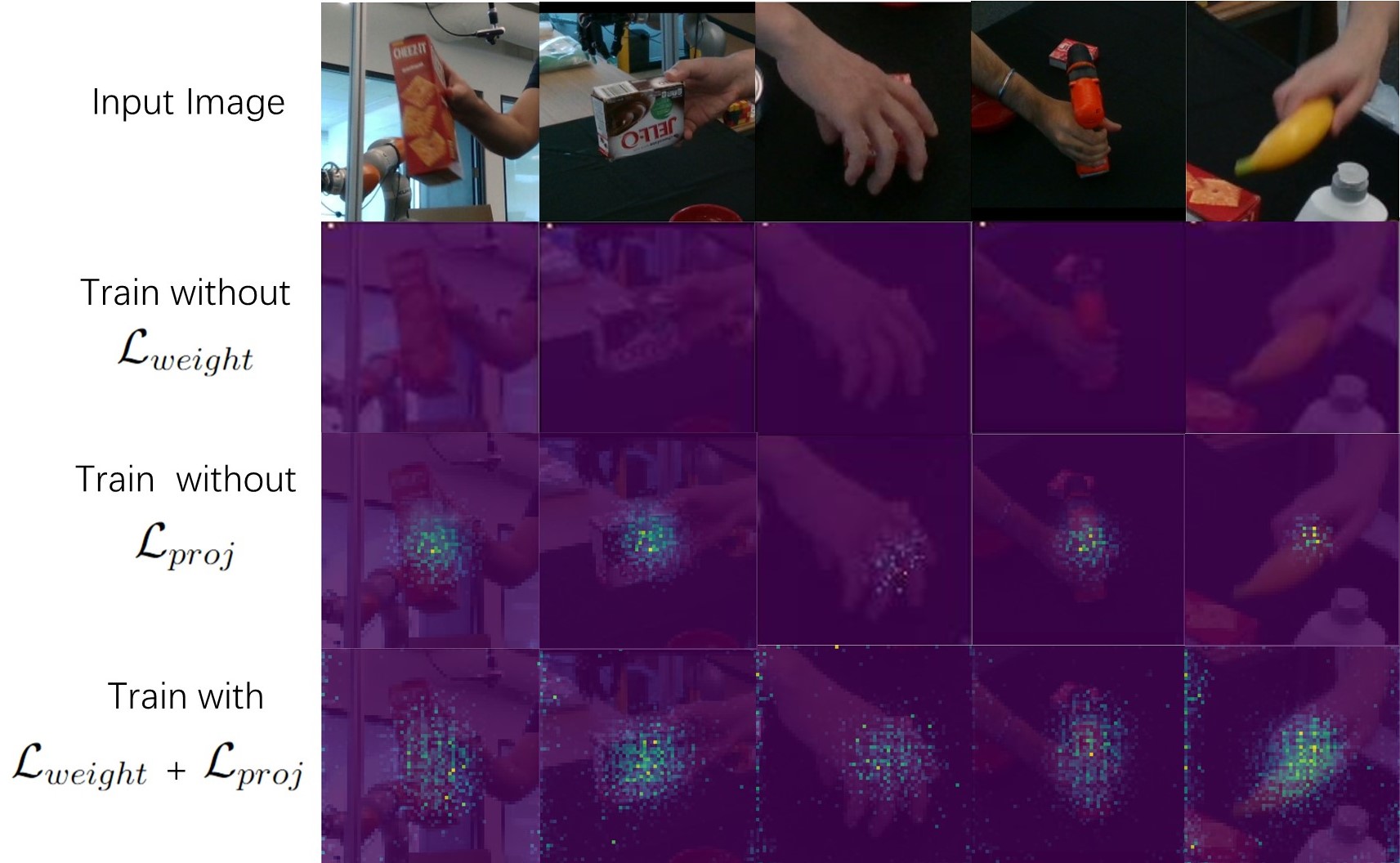}
  \caption{Attention maps of 512 query points on the shape prior. The bright areas indicate the high probability to be an object region.}
  \label{fig12}
  \vspace{-.15in}
\end{wrapfigure}

\textbf{Effect of two vision-guided losses.} 
We introduce two vision-guided loss terms, \ie, $\mathcal{L}_{weight}$ and $\mathcal{L}_{proj}$, to regulate attention mechanisms in correlating 3D prior patches with 2D image regions. 
Our ablation studies on the DexYCB dataset (Table \ref{table:seven}) validate the effectiveness of $\mathcal{L}_{weight}$ and $\mathcal{L}_{proj}$. When training without $\mathcal{L}_{weight}$, we observe a significant increase in the median Chamfer distance between the centered predicted point clouds and ground truth. Similarly, removing $\mathcal{L}_{proj}$ leads to a 0.7 increase in this metric. We further visualize attention maps in Figure \ref{fig12} for 512 points with and without these two losses. 
Without $\mathcal{L}_{weight}$, all query points incorrectly focus on the zero (u,v) region, forcing the decoder to use uninformative top-left corner features for coordinate prediction. 
Without $\mathcal{L}_{proj}$, attention becomes overly concentrated at the object center, neglecting edge features. 
Joint application of both losses enables spatially distributed attention, providing the decoder with comprehensive local visual cues for decoding.


\noindent\textbf{Limitation.}  TIGeR inherits the constraints from off-the-shelf generative models. Specifically, our current implementation struggles with objects exhibiting significant intra-class shape diversity under functional states, \eg, modeling both open and closed configurations of scissors. This limitation stems from existing generative priors prioritizing inter-class discriminability over fine-grained state variations, occasionally leading to ambiguous geometric reconstructions in dynamic interaction scenarios. Based on more future work on fine-grained 3D generation, our method would further improve the scalability.

\section{Conclusion}
In this paper, we present a new approach, Text-Instructed Generation and Refinement (TIGeR), for 3D hand-object interaction estimation that addresses the scalability of template-based approaches.  
By synergizing cross-modal generative model with geometric refinement, TIGeR eliminates reliance on hand-crafted templates while maintaining interpretability through its two-stage design, \ie, text-instructed prior generation and vision-guided refinement. We also provide a cookbook to complete prior registration and shape deformation using attention blocks to fuse the local 2D visual features and 3D geometric features. 
Extensive evaluations on two widely-used benchmarks, \ie, DexYCB and Obman, verify the effectiveness of the generated 3D prior, outperforming existing methods by 0.034 in F-score@5 while reducing shape reconstruction Chamfer Distance by 0.69. The framework's generalizability is further evidenced by its robustness against occlusions and seamless integration with different priors, \eg, retrieved priors.  

%
%
{\small
\bibliographystyle{splncs04}
\bibliography{neurips}
}

\newpage

\appendix

\clearpage
\section*{NeurIPS Paper Checklist}

\begin{enumerate}

\item {\bf Claims}
    \item[] Question: Do the main claims made in the abstract and introduction accurately reflect the paper's contributions and scope?
    \item[] Answer: \answerYes{} 
    \item[] Justification: We have introduced our motivation, contribution, and method both in the abstract and introduction.

\item {\bf Limitations}
    \item[] Question: Does the paper discuss the limitations of the work performed by the authors?
    \item[] Answer: \answerYes{} 
    \item[] Justification: TIGeR inherits the constraints from off-the-shelf generative models. Specifically, our current implementation struggles with objects exhibiting significant intra-class shape diversity under functional states, \eg, modeling both open and closed configurations of scissors. This limitation stems from existing generative priors prioritizing inter-class discriminability over fine-grained state variations, occasionally leading to ambiguous geometric reconstructions in dynamic interaction scenarios. Based on more future work on fine-grained 3D generation, our method would further improve the scalability.

\item {\bf Theory Assumptions and Proofs}
    \item[] Question: For each theoretical result, does the paper provide the full set of assumptions and a complete (and correct) proof?
    \item[] Answer: \answerNA{} 
    \item[] Justification: We do not include theoretical results. 

    \item {\bf Experimental Result Reproducibility}
    \item[] Question: Does the paper fully disclose all the information needed to reproduce the main experimental results of the paper to the extent that it affects the main claims and/or conclusions of the paper (regardless of whether the code and data are provided or not)?
    \item[] Answer: \answerYes{} 
    \item[] Justification: We will release our codes soon. Please also refer to the implementation details. 

\item {\bf Open access to data and code}
    \item[] Question: Does the paper provide open access to the data and code, with sufficient instructions to faithfully reproduce the main experimental results, as described in supplemental material?
    \item[] Answer: \answerNA{} 
    \item[] Justification: We will release our codes soon.

\item {\bf Experimental Setting/Details}
    \item[] Question: Does the paper specify all the training and test details (\eg, data splits, hyperparameters, how they were chosen, type of optimizer, etc.) necessary to understand the results?
    \item[] Answer: \answerYes{} 
    \item[] Justification: We have specified training details in implementation details section, which includes optimizer setting and resources demands. Noting we follow previous works for some chosen of hyper-parameter for a fair comparison. 

\item {\bf Experiment Statistical Significance}
    \item[] Question: Does the paper report error bars suitably and correctly defined or other appropriate information about the statistical significance of the experiments?
    \item[] Answer: \answerNA{} 
    \item[] Justification: We mainly follow the previous work to report the evaluation metric. We report the average result with 5 runs. 

\item {\bf Experiments Compute Resources}
    \item[] Question: For each experiment, does the paper provide sufficient information on the computer resources (type of compute workers, memory, time of execution) needed to reproduce the experiments?
    \item[] Answer: \answerYes{} 
    \item[] Justification: The computation resources specific is recorded in implementation details section.
    
\item {\bf Code Of Ethics}
    \item[] Question: Does the research conducted in the paper conform, in every respect, with the NeurIPS Code of Ethics \url{https://neurips.cc/public/EthicsGuidelines}?
    \item[] Answer: \answerYes{} 
    \item[] Justification: We carefully follow the NeurIPS code of Ethics. 

\item {\bf Broader Impacts}
    \item[] Question: Does the paper discuss both potential positive societal impacts and negative societal impacts of the work performed?
    \item[] Answer: \answerYes{} 
    \item[] Justification: Please see the Appendix. 
    
\item {\bf Safeguards}
    \item[] Question: Does the paper describe safeguards that have been put in place for responsible release of data or models that have a high risk for misuse (\eg, pretrained language models, image generators, or scraped datasets)?
    \item[] Answer: \answerYes{} 
    \item[] Justification: Our work based on already open-sourced model and we keep the default ill-content filter on.

\item {\bf Licenses for existing assets}
    \item[] Question: Are the creators or original owners of assets (\eg, code, data, models), used in the paper, properly credited and are the license and terms of use explicitly mentioned and properly respected?
    \item[] Answer: \answerYes{}
    \item[] Justification: Our work is based on open-sourced model and dataset. We have cited all the work we used.

\item {\bf New Assets}
    \item[] Question: Are new assets introduced in the paper well documented and is the documentation provided alongside the assets?
    \item[] Answer: \answerNA{} 
    \item[] Justification: We doesn't release new assets.

\item {\bf Crowdsourcing and Research with Human Subjects}
    \item[] Question: For crowdsourcing experiments and research with human subjects, does the paper include the full text of instructions given to participants and screenshots, if applicable, as well as details about compensation (if any)? 
    \item[] Answer: \answerNA{} 
    \item[] Justification: Our work does not involve crowdsourcing nor research with human subjects.

\item {\bf Institutional Review Board (IRB) Approvals or Equivalent for Research with Human Subjects}
    \item[] Question: Does the paper describe potential risks incurred by study participants, whether such risks were disclosed to the subjects, and whether Institutional Review Board (IRB) approvals (or an equivalent approval/review based on the requirements of your country or institution) were obtained?
    \item[] Answer: \answerNA{} 
    \item[] Justification: Our work does not involve crowdsourcing nor research with human subjects.

\end{enumerate}

\section{Appendix}

\textbf{Metrics.} We evaluate the quality of both hand and object reconstruction by computing the Chamfer Distance(mm) between the predicted point clouds and the ground truth point clouds. We also report F-score at 5 mm ($FS_o@5$) and 10 mm ($FS_o@10$) as thresholds for predicted object point clouds and F-score at 1 mm ($FS_h@1$) and 5 mm ($FS_h@5$) as thresholds for predicted hand point clouds.

\textbf{Datasets.} \textbf{(1) DexYCB} contains 582,000 RGB-D frames capturing single hand grasping of objects from 8 views. It provides 3D hand pose and 6D object pose of 20 different objects from YCB-Video~\cite{ycb_video}. 
Each frame contains a target object grasped by hand and 1-3 other objects placed on a black table. 
Following~\cite{chen2023gsdf}, 
we pick out one frame from every sequential 6 frames as the training sample. Finally, 
we build the training set with 29k samples and the test set with 5k samples. We crop original 640$\times$480 RGB images into 256$\times$256 images that are centered around the target object. 
\textbf{(2) Obman} is a large-scale synthesis dataset built by Hasson \textit{et al.}~\cite{hasson19_obman}, which contains 150k images. 
They select 2772 meshes of 8 object categories from ShapeNet~\cite{shapenet2015} and use GraspIt~\cite{Graspit2004} software and MANO~\cite{mano2017} model to generate hand poses and hand meshes. The object meshes and hand meshes are rendered on the background images with the size of 256$\times$256, which are sampled from LSUN~\cite{lsun2016} and ImageNet~\cite{imagenet2009}. 
Following gSDF~\cite{chen2023gsdf}, we preprocess and split the Obman dataset into a training set with 87k samples and a test set with 6k samples.

\textbf{Compared Methods.} We compare our method with three competitive template-free methods. (1) Hasson \textit{et al.}~\cite{hasson19_obman} proposes a classic template-free method to reconstruct hand and object given an RGB image. They extract global visual features from the input image and decode the features into object mesh and hand mesh by using AtlasNet~\cite{AtlasNet2018} and Mano Layers~\cite{mano2017} respectively. (2) AlignSDF~\cite{chen2022alignsdf} is an SDF-based template-free method for the hand-object interaction task. They leverage both global visual features and pose information to decode the surface of the hand and object by using an SDF decoder. (3) gSDF~\cite{chen2023gsdf} is another SDF-based method. They extract feature maps from the given RGB image, using local visual features for reconstruction. To improve the accuracy of hand pose prediction, they first predict 3D coordinates for 21 key points from the heat map and use the Inverse Kinematics~\cite{inverse_kinematics} algorithm to estimate the hand pose.

\noindent\textbf{Broader Impact.}
Our research pushes the boundaries of 3D reconstruction technology, particularly in its applicability to real-world scenarios where unconstrained interactions are the norm. By eliminating the reliance on pre-defined 3D templates, TIGeR paves the way for more dynamic, adaptable systems capable of understanding and replicating complex human-object engagements. This has profound implications for several sectors:

Positive Impacts:
(1) Innovation Boost: Our text-instructed system enhances creativity in VR, gaming, and design, allowing for realistic, interactive experiences tailored to users.
(2) Efficiency \& Safety: Robots can better grasp and handle objects, improving automation in industries like manufacturing and healthcare, keeping workers out of harm's way.
(3) Education Uplift: Students gain from lifelike simulations that make learning hands-on, especially in challenging fields.
(4) Accessibility Wins: By advancing assistive tech, we are making digital tools more inclusive for all.

Mitigating Negative Effects: (1) Job Market Shift: While some jobs may change, we emphasize retraining and creation of new tech-focused roles to support transition.
(2) Privacy Assurance: We're mindful of privacy; clear policies and strong data protection measures will be in place. (3) Bridging the Gap: Collaborations aim to ensure our tech benefits reach everyone, narrowing any digital divide.

\end{document}